\documentclass[conference]{IEEEtran}
\IEEEoverridecommandlockouts
\usepackage{cite}
\usepackage{amsmath,amssymb,amsfonts}
\usepackage{algorithmic}
\usepackage{graphicx}
\usepackage{textcomp}
\usepackage{xcolor}
\usepackage{tabularx}
\usepackage{float}
\usepackage{CJKutf8}
\usepackage{balance}
\usepackage{url} 
\usepackage{hyperref}
\usepackage{float}
\def\BibTeX{{\rm B\kern-.05em{\sc i\kern-.025em b}\kern-.08em
    T\kern-.1667em\lower.7ex\hbox{E}\kern-.125emX}}

\usepackage{booktabs} 
\makeatletter
\newcommand{\linebreakand}{%
  \end{@IEEEauthorhalign}
  \hfill\mbox{}\par
  \mbox{}\hfill\begin{@IEEEauthorhalign}
}
\makeatother

\begin{document}

\title{Are Large Language Models Possible to Conduct Cognitive Behavioral Therapy? \\
\thanks{*Corresponding author.}
}
\author{\IEEEauthorblockN{1\textsuperscript{st} Hao Shen}
\IEEEauthorblockA{\textit{School of Information Science and} \\
\textit{Engineering, Lanzhou University}\\
Lanzhou, China \\
220220942151@lzu.edu.cn}
\and
\IEEEauthorblockN{1\textsuperscript{st} Zihan Li}
\IEEEauthorblockA{\textit{School of Information Science and} \\
\textit{Engineering, Lanzhou University}\\
Lanzhou, China \\
lizihan2020@lzu.edu.cn}
\and
\IEEEauthorblockN{2\textsuperscript{nd} Minqiang Yang*}
\IEEEauthorblockA{\textit{School of Information Science and} \\
\textit{Engineering, Lanzhou University}\\
Lanzhou, China \\
yangmq@lzu.edu.cn}
\linebreakand
\IEEEauthorblockN{3\textsuperscript{rd} Minghui Ni}
\IEEEauthorblockA{\textit{School of Information Science and} \\
\textit{Engineering, Lanzhou University}\\
Lanzhou, China \\
220220942081@lzu.edu.cn}
\and
\IEEEauthorblockN{4\textsuperscript{th} Yongfeng Tao}
\IEEEauthorblockA{\textit{School of Information Science and} \\
\textit{Engineering, Lanzhou University}\\
Lanzhou, China \\
taoyf21@lzu.edu.cn}
\and
\IEEEauthorblockN{5\textsuperscript{th} Zhengyang Yu}
\IEEEauthorblockA{\textit{School of Information Science and} \\
\textit{Engineering, Lanzhou University}\\
Lanzhou, China \\
yuzhy2020@lzu.edu.cn}
\linebreakand
\IEEEauthorblockN{6\textsuperscript{th} Weihao Zheng}
\IEEEauthorblockA{\textit{School of Information Science and} \\
\textit{Engineering, Lanzhou University}\\
Lanzhou, China \\
zhengweihao@lzu.edu.cn}
\and
\IEEEauthorblockN{7\textsuperscript{th} Chen Xu*}
\IEEEauthorblockA{\textit{School of Medical Technology,} \\
\textit{Beijing Institute of Technology}\\
Beijing, China \\
chenxu05037@bit.edu.cn}
\and
\IEEEauthorblockN{8\textsuperscript{th} Bin Hu*}
\IEEEauthorblockA{\textit{School of Information Science and} \\
\textit{Engineering, Lanzhou University}\\
Lanzhou, China \\
bh@lzu.edu.cn}
}

\maketitle

\begin{abstract}
In contemporary society, the issue of psychological health has become increasingly prominent, characterized by the diversification, complexity, and universality of mental disorders. Cognitive Behavioral Therapy (CBT), currently the most influential and clinically effective psychological treatment method with no side effects, has limited coverage and poor quality in most countries. In recent years, researches on the recognition and intervention of emotional disorders using large language models (LLMs) have been validated, providing new possibilities for psychological assistance therapy. However, are LLMs truly possible to conduct cognitive behavioral therapy? Many concerns have been raised by mental health experts regarding the use of LLMs for therapy. Seeking to answer this question, we collected real CBT corpus from online video websites,  designed and conducted a targeted automatic evaluation framework involving the evaluation of emotion tendency of generated text, structured dialogue pattern and proactive inquiry ability. For emotion tendency, we calculate the emotion tendency score of the CBT dialogue text generated by each model. For structured dialogue pattern, we use a diverse range of automatic evaluation metrics to compare speaking style, the ability to maintain consistency of topic and the use of technology in CBT between different models . As for inquiring to guide the patient, we utilize PQA (Proactive Questioning Ability) metric. We also evaluated the CBT ability of the LLM after integrating a CBT knowledge base to explore the help of introducing additional knowledge to enhance the model's CBT counseling ability. Four LLM variants with excellent performance on natural language processing are evaluated, and the experimental result shows the great potential of LLMs in psychological counseling realm, especially after combining with other technological means.
\end{abstract}

\begin{IEEEkeywords}
large language models, cognitive behavioral therapy, knowledge base, evaluation metrics
\end{IEEEkeywords}

\section{Introduction}

Psychological disorders, especially emotional disorders such as depression, have become a global public health issue. According to the latest statistics from the World Health Organization (WHO), about 280 million people worldwide suffer from depression, more than 700,000 people commit suicide due to depression attacks worldwide, and suicide has become the fourth leading cause of death among people aged 15–29 \cite{depression2023who}. Depression has a significant impact on an individual's quality of life, affecting all aspects of their life and leading to varying degrees of functional impairment, including but not limited to impaired social function, decreased work ability, sleep disorders, mental fatigue, and increased potential risk of suicide \cite{kessler2005lifetime}. Therefore, effective treatment and intervention of psychological disorders have become a global focus of attention. 

At present, the treatment for mental illness is usually divided into drug therapy and non-drug therapy. Drug therapy often allows patients with mental illness to take corresponding antipsychotic drugs, which can inhibit the patient's condition immediately after taking them, but long-term dependence on antipsychotic drugs is often accompanied by many side effects, including but not limited to memory loss, dizziness and fatigue, cognitive impairment \cite{bandelow2020current}. Therefore, psychological intervention is often more applicable. In order to alleviate the emotional intensity of patients, ordinary psychological counseling with empathetic characteristic usually says very general words, such as "don't worry, you can overcome difficulties".
Although this kind of conversation can indeed alleviate the patient's negative emotions in the present, it does not fundamentally solve the mental problem of the patient.
By contrast, CBT, as a simple, fast, non-invasive, and easy to operate structured psychological counseling technique, regards cognition as breakthrough point and attempts to fundamentally help the patient to get rid of mental disorders \cite {katz2020excessive, marker2019intermittent, olsson2021treatment}. 

In CBT perspective, patients with mental illness often have cognitive distortion. And CBT technology takes the interpretation of cognitive influence on emotion as the core concept. By adjusting patients' unreasonable thinking patterns and abnormal cognition of people or events, the problem behaviors are adjusted successively to reduce the symptoms of patients. For example, a student faces the exam (event), worries about the exam result is not satisfactory (cognition), and thus appears anxious (emotion). In this case, a CBT therapist may uses technologies like socratic questioning and listing objective evidence to challenge the incorrect cognition, thereby alleviating the student's anxious mood.

Although psychological counseling and psychotherapy are widely regarded as effective treatment methods, there are a series of challenges and drawbacks at present. For example, the lack of investment in mental health care in many regions, the lack of resources and the high cost of face-to-face psychotherapy, has resulted in a large number of patients not getting the help they need in a timely manner. In addition, face-to-face psychological counseling also has problems such as personal privacy disclosure, social discrimination against patients with mental illness, and unstable treatment effects, which limit its potential application in large-scale mental health intervention. Therefore, there is an urgent need to explore new technological means and treatment models to improve the quality and accessibility of mental health services.

The application of AI technology in detecting psychological disorders has become very prevalent and effective \cite{yang2023trial, tao2024depmstat, tao2024depressive, shen2019improved}. In recent years, the rise of LLMs has prompted researchers to explore their potential applications in the field of mental health \cite{tao2023classifying}. LLMs are computational models with the ability to understand and generate human language, which can predict the likelihood of word sequences or generate new text based on given input \cite{gao2004introduction}, which makes them useful for virtual psychotherapists to have conversations with patients, assess and screen patients for mental health issues, and provide timely interventions and recommendations. Therefore, the study of psychological facilitation ability based on LLMs can combine the diagnosis and treatment principle of traditional CBT technology with the advantages of high efficiency of information processing, analysis and response generation of LLM, so as to generate more targeted treatment content and achieve the same or even better treatment effect as traditional CBT. The process of utilizing LLM-based CBT therapist for assisted psychological counseling can also be combined with the detection of mental disorders, such as depression detection \cite{hu2023detecting, yang2023three, yang2022clustering}. Moreover, Yang et al. \cite{yang2024behavioral} proposed to combine behavioral information feedback with LLMs by utilizing additional universal and inexpensive devices such as eye tracker \cite{yang2023wearable} to improve the CBT ability of LLMs. 

The psychological counseling based on the LLM can break through the limitations of time and space, provide more flexible treatment methods, improve the degree of individuation and effect of treatment, reduce the cost and threshold of patient treatment, and reduce the pressure and stigma of some patients for mental illness treatment, so as to improve their mental health. However, concerns have been raised by mental health experts regarding the use of LLMs for therapy as the therapy provided may not be accurate \cite{moon2024caremultimodaldepressiondetection}. To provide treatment for people with mental illness in a more standardized manner, it is of great practical significance to conduct comprehensive assessment to improve the level of virtual mental health diagnosis and treatment. Unlike previous generated CBT dialogue datasets, our dataset contains conversations that reach up to hundreds of turns and conversations with more complex situations, which poses difficulties for human evaluation. On the other hand, we can better simulate long CBT counseling sessions to evaluate LLMs more accurately by using the real CBT dialogue dataset, because the real CBT sessions are naturally structured.
Moreover, existing methods seldom comprehensively consider the specific attributes of CBT. Therefore, this paper proposed a comprehensive automatic evaluation framework to assess the CBT counseling ability of the LLM, which includes evaluation of emotion tendency, structured dialogue pattern and inquiry ability. During the experiment, we found that most metrics of the general LLMs are very unsatisfactory. Thus, we tried to introduce an additional CBT knowledge base and evaluate the degree to which knowledge base integration improves the CBT ability of the LLM.

The main contributions of this paper are as follows:

$\bullet$ We designed a comprehensive automatic evaluation framework for evaluating CBT ability of the LLM, including evaluation of emotion tendency, structured dialogue pattern and inquiry ability.

$\bullet$ In addition to evaluating the general LLMs, we constructed a CBT-related knowledge base and integrated it with the general models to explore whether introducing additional knowledge can enhance the CBT capability of the models.

The rest of this paper is structured as follows: Section \ref{Related} discusses related works; Section \ref{Methods} explains the CBT dataset, LLMs to evaluate, and the evaluation framework; Section \ref{Results} demonstrates the outcomes of the evaluation and makes a detailed analysis; and Section \ref{Conclusion} summarizes the work of this paper and points out some limitations.

\section{Related Works} \label{Related}

Nowadays, the application of LLMs in psychotherapy is still in early stage, but there have been some studies exploring its potential value. In recent years, natural language processing technology has been applied in the development of mental health applications. Chatbots that incorporate psychotherapy theories, such as Woebot \cite{prochaska2021therapeutic} and Wysa \cite{inkster2018empathy}, have demonstrated significant therapeutic benefits. They can alleviate symptoms and engage in effective psychotherapy conversations with patients. Additionally, open-source models in psychology have emerged, such as SoulChat \cite{chen2023soulchat} and MeChat \cite{qiu2023smile}. These models can offer emotional support more akin to face-to-face psychotherapy by enhancing the model's empathetic capacity, guiding users to express themselves, and offering sound advice. The emergence of these applications signifies an innovative trending in mental health support realm, since natural language processing technology is expected to increase the accessibility and efficiency of mental health services. 

Although LLMs have achieved remarkable success, they still have obvious limitations such as producing hallucinations, especially in domain-specific or knowledge-intensive tasks \cite{kandpal2023large, zhang2023siren}.
To fully leverage the capability of the LLM in a specific domain, it is a good choice to integrate external knowledge bases with the model itself. 
This integration of knowledge base and LLM is called RAG (Retrieval-Augmented Generation) \cite{zhao2024retrieval}, which serves to enhance the model's language understanding and generation abilities, enabling it to provide precise and concise responses to natural language queries. 
A typical RAG process involves indexing, retrieval, and generation. In indexing step, knowledge documents are split into chunks, encoded into vectors, and stored in a vector database; in retrieval step, the Top k document chunks most relevant to the question based on semantic similarity are retrieved; and in generation step, original input and the retrieved chunks are concatenated and then input into the LLM for answer generating \cite{gao2023retrieval}.
Domain-specific knowledge bases can provide crucial data and information support for intelligent applications and decision-making, thereby promoting research and innovation within the corresponding field.

As LLMs continue to play a crucial role in both research and daily life, the evaluation of these models is increasingly important as it serves as a fundamental basis for enhancing the performance of understanding and generating text. The current evaluation metrics can be divided into the following categories \cite{chang2024survey, bommasani2023holistic}:

$\bullet$ Traditional evaluation metrcis. In the early stage, the evaluation of small-scale language models mainly used traditional evaluation metrics, such as Precision, Accuracy, Recall, etc. These metrics can reflect the prediction accuracy, generalization ability, and applicability of the model.

$\bullet$ Task-based evaluation metrics. As language models are applied to various natural language processing tasks, task-based evaluation metrics have become more important. For example, commonly used evaluation metrics in machine translation tasks include BLEU, METEOR, etc., while metrics such as ROUGE may be employed for text generation tasks.

$\bullet$ Human evaluation methods. Manual evaluation mainly involves five aspects: types of evaluators, evaluation scales, providing references and context, absolute/relative evaluation, and evaluators providing necessary explanations for their own decisions. No matter what kind of configuration is used, there will always be multiple results from different people for an output, which will always be integrated through various methods to obtain a final score.

$\bullet$ Improved automatic evaluation metrics. Automatic evaluation metrics such as BLEU are mainly based on $n$-gram overlap for evaluation, and their perception ability of semantic information is weak. Improved automatic evaluation metrics aim to enhance the measurement of models' understanding of text semantics, more accurately reflecting the quality of the text. By synthesizing multi-dimensional elements like semantic similarity, grammatical correctness, and diversity, improved automatic evaluation metrics can better evaluate the performance of LLMs, providing more accurate references for model training and optimization.

$\bullet$ Robustness, ethics, bias, and trustworthiness evaluation. Robustness evaluation includes evaluating the model's resistance to adversarial sample attacks, its performance under abnormal input conditions, and its ability to handle noise, ambiguity, and other issues; Ethics evaluation includes evaluating whether the text generated by the model contains harmful, illusory or discriminatory content; Bias evaluation includes evaluating whether there is bias or discrimination in gender, race, region and other aspects in the text generated by the model; Trustworthiness evaluation encompasses evaluating the accuracy, truthfulness, and alignment with real-world standards of the content generated by the model, as well as the user's trust and acceptance of the generated text.

\section{Methods}\label{Methods}

\subsection{Dataset and Large Language Model}

The video files of therapists' CBT sessions with patients utilized in this paper come from public social media. There are 46 Chinese conversation files and 150 English conversation files, totaling 6386 turns of dialogue. Some of the conversations in our dataset reach up to hundreds of turns. And we translate English dialogues into Chinese dialogues for evaluation. All conversations represent authentic treatment scenarios or role-played simulated clinical cases encompassing treatment processes of various disorders and issues, such as anxiety, depression, post-traumatic stress disorder, and obsessive-compulsive disorder. Some of dialogues are presented in Table \ref{dataset-example}.

\begin{table}[htbp]
    \centering
    \renewcommand{\arraystretch}{1}
    \caption{Examples of the CBT dataset}
    \begin{tabularx}{0.47\textwidth}{lX}
        \toprule
        \multicolumn{1}{c}{Role} & \multicolumn{1}{c}{Content} \\
        \midrule
        \multicolumn{2}{c}{Scene: Social barrier} \\
        \midrule
        therapist & You said the first time you felt anxious was five or six years ago, when you were about 19 years old, right? \\
        \midrule
        patient & Yeah. \\
        \midrule
        therapist & What happened at that time? \\
        \midrule
        patient & I was a bit shy when I was at school, but I don't think it had a significant impact on things like in the past few years. But I refuse to attend the party or refuse invitations. I think my job can help me because I love photography. I really enjoy photography, but over time, my work has become less and less. So this is my fault, I keep saying no. \\
        \midrule
        therapist & Which job is the most difficult for you? \\
        \midrule
        patient & Well, I want to work with others when I have to. I don't like that. Just like when I try to do an event or something with many people there, it's not just a photographer. If I think of taking photos of people I don't know, I find myself rejecting it. I mean, I want to do it, but it's too scary, you know, so I don't do it. \\
        \midrule
        therapist & When asked to do a job that involves taking photos with a group of people, you need to have social interaction. What's on your mind? \\
        \midrule
        patient & I feel really hot and sweaty, I feel like everyone is watching me, I feel like I'm really stupid, I can't complete my work, I'm just an idiot. You know, what I mean is, they will think I look like a weirdo. \\
        \bottomrule

        \multicolumn{2}{c}{Scene: Anxiety} \\
        \midrule
        patient & I have been reflecting on a lot lately. The company has been working overtime and very busy lately, and we are going to work on a new project. This project has given me a headache, and I haven't been able to sleep well lately. \\
        \midrule
        therapist & Some of the company's projects need to be completed, causing you a lot of trouble. \\
        \midrule
        patient & Yeah, on my way here by bike today, a big truck drove by. I suddenly felt very nervous and unconsciously imagined what would happen if the truck overturned and ran over me, which made me very scared. I don't know if it's related to my previous work experience. \\
        \midrule
        therapist & Since you first thought of such a question, why don't we start by talking about a truck that seems to be coming over? Have you ever had such an idea in your mind before? \\
        \midrule
        patient & Previously, there were also some, occasionally some. At that time, I thought it was quite normal, and people might be afraid of death. But this time, it may be because of the recent high pressure, and I came here today before consulting. This feeling will be stronger, which makes me a bit scared.\\
        \midrule
        therapist & What do you think the odds are that the truck might flip over and run over you? \\
        \midrule
        patient & Below 10\%, it feels relatively low. \\
        \bottomrule
    \end{tabularx}
    \label{dataset-example}
\end{table}

As for LLMs, we choose ChatGPT-3.5-turbo, ERNIE-3.5-8K,  iFlytek Spark V3.0 and ChatGLM-3-turbo, because they have leading performance and convenient APIs. This paper focuses on these four representative LLMs to evaluate and compare their capability of conducting CBT conversations.

\subsection{Evaluation Framework and Evaluation Metric}

\begin{figure*}[htbp]
\centering 
\includegraphics[scale=0.6]{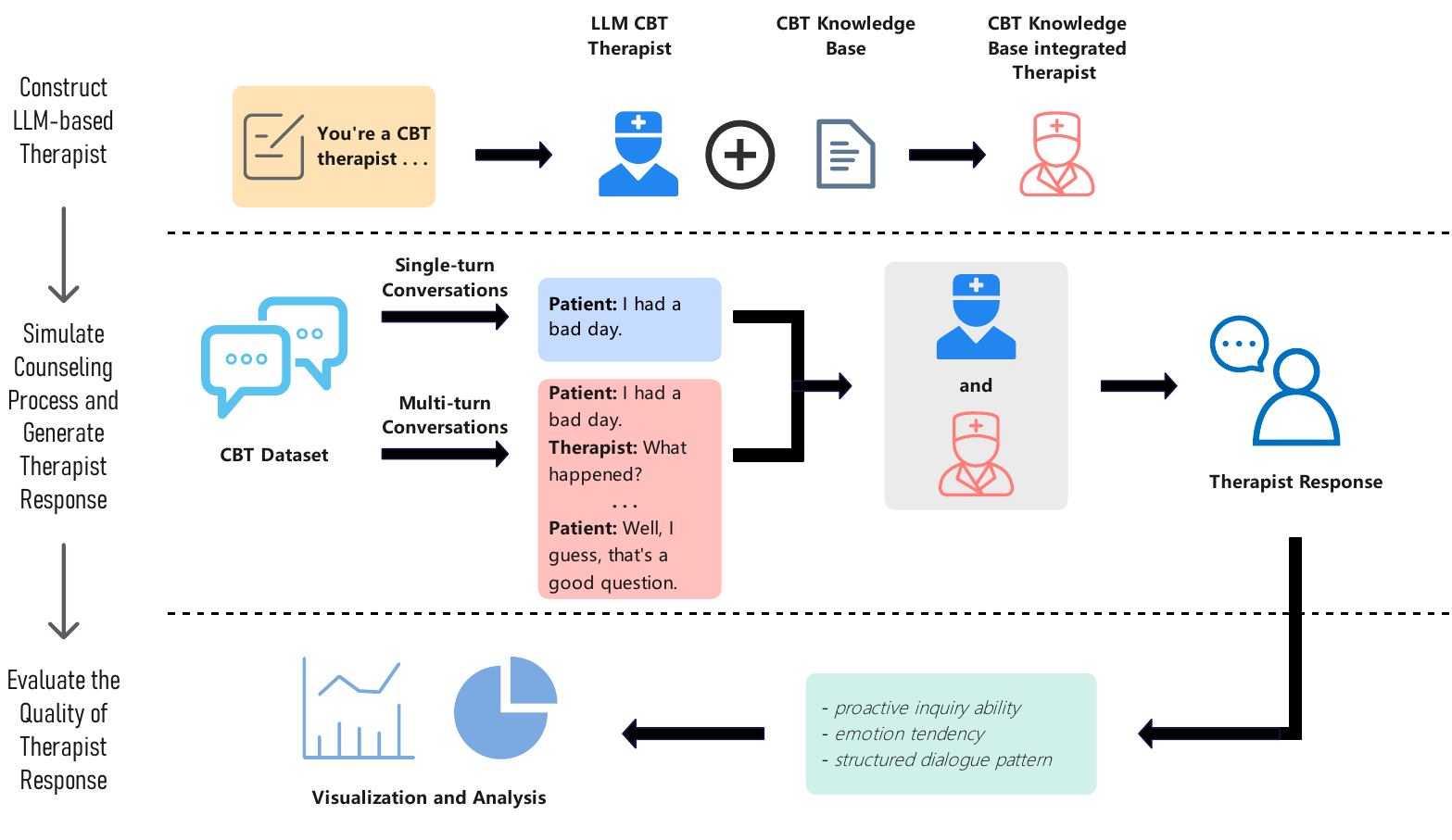} 
\caption{The overview of our automatic computational evaluation framework. We constructed two types of LLM-based CBT therapists and assessed their CBT counseling ability under single-turn conversations and multi-turn conversations.}
\label{framework}
\end{figure*}

CBT is a complex and persistent process, therefore the ability of LLMs to understand and remember conversation contexts in different interaction situations should be considered. Our proposed computational evaluation framework considers two usage scenarios, each consisting of the same three aspects. As shown in Fig. \ref{framework}, we evaluate the CBT ability of the general LLMs and the CBT knowledge base integrated LLMs in terms of emotion tendency, structured dialogue pattern and inquiry ability under single-turn conversation and multi-turn conversation. Single-turn conversations focus on extracting key information from current input and offering real-time appropriate feedback; multi-turn conversations focus more on topic consistency, the ability to communicate continuously. And under multi-turn conversation, models take 5 rounds of dialogue as one input. The integrated database contains books on CBT technology, as well as a small portion of the CBT dialogue dataset collected from the internet. These CBT knowledge texts are processed in a structured manner and then uploaded through plugin interfaces.

Given the intrinsic nature of mental health counseling, it is essential to restrict the emotion tendency of models' responses, as negative emotions could have severe consequences for vulnerable patients. And in the actual CBT process, psychotherapists frequently employ a large number of questions to accurately gather detailed information from the patient. Therefore, the ability of LLMs to proactively inquire and guide the patient is also crucial. Besides, considering the structured characteristic of CBT, we also evaluate the structured dialogue pattern including the speaking style, the ability to maintain consistency of topic and the use of technology in CBT. 

There have been some studies on evaluating the proactive inquiry ability of the LLM \cite{chen2023bianque, deng2023knowledge}. we use PQA to perform quantization calculations because it can intuitively reflect the proportion of inquiries in all responses. For emotion tendency, we directly calculate the emotion tendency score. To visually reflect speaking style of the LLMs, word cloud maps under single-turn and multi-turn conversations are drawn and compared. Meanwhile, we can observe whether knowledge base integration can help the LLM use CBT techniques through word cloud maps. Benefiting from the collected real CBT sessions that are naturally structured, automatic metrics like metrics on semantic similarity are adequate reflect the structured dialogue capability of the LLMs. And since human evaluation has lower efficiency and higher cost when facing conversations that can reach up to hundreds of turns, various automatic evaluation metrics are used to evaluate models' ability to maintain consistency of topic and the use of technology in CBT. Depending on whether rely on pre-trained models or large-scale datasets, context-free metrics can be roughly divided into two categories: untrained metrics and trained metrics \cite{sai2022survey}. The untrained context-free metrics used in this paper include DISTINCT \cite{li2015diversity}, ROUGE \cite{chin2004rouge}, METEOR \cite{banerjee2005meteor}, chrF \cite{popovic2015chrf}, chrF++ \cite{popovic2017chrf++} and BERTScore \cite{zhang2019bertscore}, and the trained context-free metrics include BLEURT \cite{sellam2020bleurt} and BartScore \cite{yuan2021bartscore}. By conducting these two types of evaluation metrics simultaneously, the structured dialogue pattern of each LLM can be more objectively and comprehensively evaluated.

\section{Results} \label{Results}

\subsection{General Large Language Models}

In a single-turn conversation, LLMs only consider the current input text without contextual constraints, so it may be more inclined to use open generation, which will generate more text as replies. In contrast, in multi-turn conversations, LLMs need to consider the previous conversation history, which is limited by the context, and therefore tend to use targeted generation, resulting in a relatively small amount of generated text. Table \ref{tab-single} and Table \ref{tab-multi} separately show a response example of each LLM under single-turn conversations and multi-turn conversations.

\begin{table}[htbp]
    \centering
    \renewcommand{\arraystretch}{1}
    \caption{Examples of generated single-turn CBT text of each general large language model}
    \begin{tabular}{p{0.09\textwidth}p{0.32\textwidth}}
        \toprule
        \multicolumn{1}{c}{Name} & \multicolumn{1}{c}{Content} \\
        \midrule
        Input & Well, I really try to make sure I take at least one shower every day, but I often find that I haven't completed the other household chores that need to be done. There are many plates and things at home that I haven't cleaned up yet. \\
        \midrule
        Reference & It makes sense that when you have this feeling, you don't have the energy to do those things. Are there any other things you used to do frequently but don't do much now? \\
        \midrule
        ChatGPT & I would like to ask why you think you can persist in taking a shower every day, but find it difficult to complete other household tasks? What do you think is the difference between the two? \\
        \midrule
        ERNIE Bot & Household chores do require a certain amount of time and effort to keep the house tidy and organized. If you find that you are still unable to complete household chores, you can consider seeking help from family or friends, or hiring a domestic service to help you share the burden of household chores. \\
        \midrule
        iFlytek Spark & This sounds like you may need to better manage your time and household tasks. Here are some helpful suggestions: develop a plan, break down tasks, and set priorities. Remember, everyone has busy times, and sometimes you may not be able to complete all the household chores, which is normal. Don't be too harsh on yourself. \\
        \midrule
        Zhipu GLM & Household chores are indeed a necessary and tedious task, especially in busy or fast-paced environments. Remember, the most important thing is to find a suitable pace of life, maintain a balance between household chores and self-care, and not neglect your health and happiness because of household chores. \\
        \bottomrule
    \end{tabular}
    \label{tab-single}
\end{table}

\begin{table}[htbp]
    \centering
    \renewcommand{\arraystretch}{1}
    \caption{Examples of generated multi-turn CBT text of each general large language model}
    \begin{tabular}{p{0.09\textwidth}p{0.32\textwidth}}
        \toprule
        \multicolumn{1}{c}{Name} & \multicolumn{1}{c}{Content} \\
        \midrule
        Input & Well, I really try to make sure I take at least one shower every day, but I often find that I haven't completed the other household chores that need to be done. There are many plates and things at home that I haven't cleaned up yet. \\
        \midrule
        Reference & It makes sense that when you have this feeling, you don't have the energy to do those things. Are there any other things you used to do frequently but don't do much now? \\
        \midrule
        ChatGPT & It sounds like you also have difficulties in self-care in your daily life. So, besides these daily tasks, what other challenges do depression bring? \\
        \midrule
        ERNIE Bot & It sounds like you've done your best, that's great. So have you tried to find some activities that make you feel happy or relaxed? \\
        \midrule
        iFlytek Spark & Depression can affect normal lifestyle habits, and sometimes a combination of medication and psychotherapy can help alleviate symptoms of depression. \\
        \midrule
        Zhipu GLM & Depression can indeed change a person's daily habits. If you feel the situation is severe or your symptoms of depression persist, it is strongly recommended that you consult a doctor or mental health specialist. \\
        \bottomrule
    \end{tabular}
    \label{tab-multi}
\end{table}

\begin{figure*}[htbp]
\centering 
\includegraphics[scale=0.25]{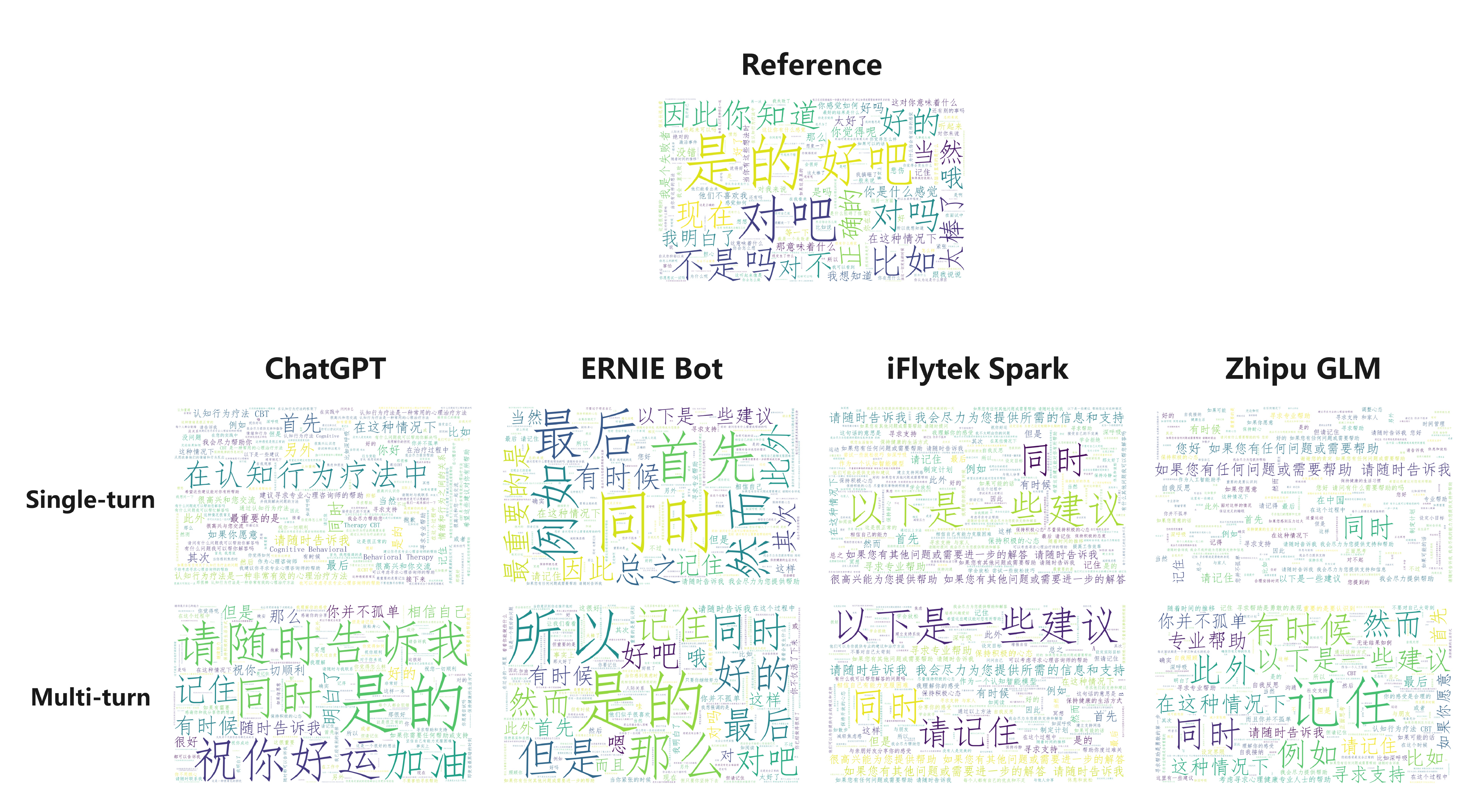} 
\caption{Cloud maps of the reference text and the text generated by general large language models.}
\label{cloudmap-general}
\end{figure*}

In order to efficiently grasp the verbal expressions commonly utilized by human psychotherapists in the communication process and the primary topics of concern during CBT, word cloud maps (Fig. \ref{cloudmap-general}) are created to visually represent the frequency of words in the text data obtained from CBT dialogue text generated by various LLMs after single-turn and multi-turn conversations.

According to the word cloud map of the reference text, it is evident that whether engaging in a single-turn or multi-turn conversations, human psychological counselors consistently demonstrate strong empathy and consider the emotions of their patients. As a result, they tend to utilize more compassionate language and gently encouraging words. 
For instance, affirming expressions like "right" hold a prominent position in the CBT reference text, while casual terms such as "well" and "yeah" are frequently used. 
These findings indicate that human psychological therapists adeptly employ gentle language to guide and comfort patients' heightened emotions.

By analyzing the CBT text generated by LLMs, it is evident that the text data generated for single-turn conversations is very different from that generated for multi-turn conversations. In single-turn conversations, prominent theme words with high frequency are only present in the ERNIE Bot's generated text set, indicating its closer resemblance to real-life situations among the four models. This can be attributed to the limitations imposed on text generation by LLMs due to dialogue content constraints, resulting in more diverse responses compared to multi-turn conversations which provide more comprehensive information. In multi-turn conversations, key words in the word cloud of each LLM's generated text are more prominent than those in single-turn conversations due to their contextual relevance and coherence. Additionally, words such as "yes", "so", "right" and "sometimes" appear more frequently in word clouds generated from multi-turn conversations, reflecting the suitability of LLMs for CBT processes involving multi-turn conversations.

\begin{figure*}[htbp]
\centering 
\includegraphics[scale=0.4]{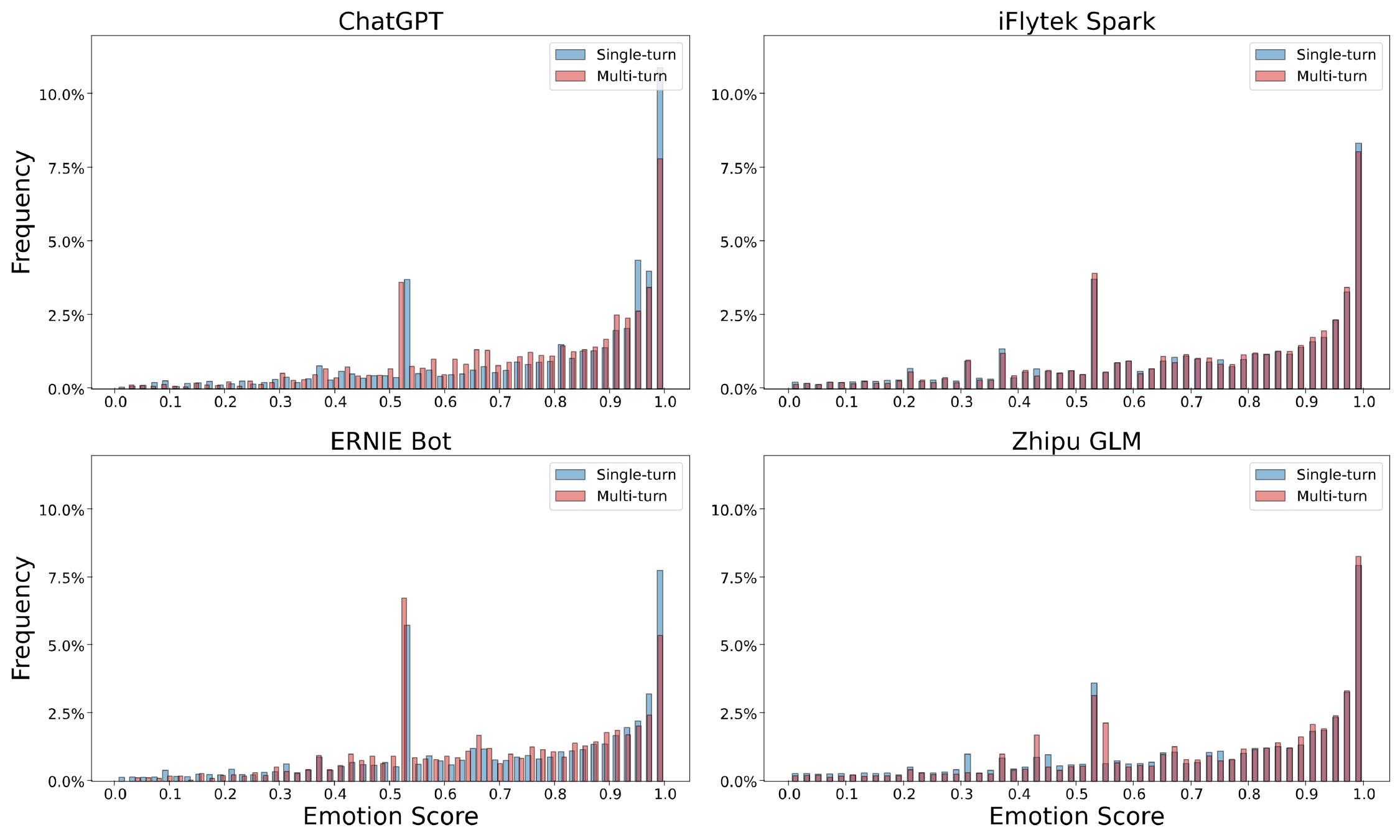} 
\caption{Emotion Scores of the CBT text generated by different general large language models. Blue bars represent the emotion score distribution under single-turn conversation, and the red bars represent the emotion score distribution under multi-turn conversation.}
\label{emotion-general}
\end{figure*}

We further conducted sentiment analysis on the CBT dialogue text generated by LLMs in both single-turn and multi-turn conversations, and the corresponding emotional tendency score was obtained (Fig. \ref{emotion-general}). Obviously, compared with the distribution of emotion scores in CBT conversations conducted by human psychological counselors, the text generated by different language models showed more positive emotions, especially the text with an emotion score of 1, which means that models generated more replies containing extremely positive emotions. This may be due to the fact that generating negative content can trigger discomfort or even psychological harm for patients, so models are often designed to be more geared towards generating positive content to reduce these risks. In summary, CBT conversation text generated by LLMs tended to contain higher positive emotions to promote continued use and increase patients' trust in LLMs.

To objectively evaluate the CBT conversational ability of each model, we utilized various standard evaluation metrics. As shown in Table \ref{metric-single-general} and Table \ref{metric-multi-general}, in addition to the DISTINCT Rate, models' ability to generate multi-turn CBT text was significantly improved in each metric. Multi-turn conversations could effectively improve the models' ability to generate CBT dialogues, especially in terms of the four models' understanding of text input, semantic similarity, and fluency of text generation. Most models experience a decrease in DISTINCT Rate during multi-turn conversations, which is actually a good sign in CBT ability evaluation. This is attributed to the fact that models take into account the history and context of previous dialogues in order to ensure coherence in their responses. And models tend to either continue with the previous topic or generate responses within similar topics, leading to a reduction in text diversity. 

\begin{table}[htbp]
\centering
\setlength{\tabcolsep}{3pt}
\caption{Evaluation metrics of general large language models under single-turn conversations}
\begin{tabular}{ccccc}
\toprule
 & ChatGPT & ERNIE Bot & iFlytek Spark & Zhipu GLM \\
\midrule
DISTINCT Rate & 0.9429 & 0.9408 & 1.0048 & 0.9779 \\
ROUGE & 0.1023 & 0.0755 & 0.0835 & 0.0748 \\
METEOR & 0.0822 & 0.0492 & 0.0732 & 0.0706 \\
chrF & 0.0551 & 0.05188 & 0.0479 & 0.0452 \\
chrF++ & 12.8161 & 11.9992 & 11.0307 & 11.2827 \\
BERTScore & 0.6486 & 0.6236 & 0.6315 & 0.6262 \\
BartScore & -4.3118 & -4.2590 & -4.3350 & -4.3507 \\
BLEURT & 0.6888 & 0.6034 & 0.6318 & 0.5554 \\
PQA & 0.6714 & 0.3005 & 0.1548 & 0.2110 \\
\bottomrule
\end{tabular}
\label{metric-single-general}
\end{table}

\begin{table}[htbp]
\centering
\setlength{\tabcolsep}{3pt}
\caption{Evaluation metrics of general large language models under multi-turn conversations}
\begin{tabular}{ccccc}
\toprule
 & ChatGPT & ERNIE Bot & iFlytek Spark & Zhipu GLM \\
\midrule
DISTINCT Rate & 0.9131 & 0.9667 & 0.9814 & 0.8861 \\
ROUGE & 0.1276 & 0.1689 & 0.1135 & 0.0758 \\
METEOR & 0.0858 & 0.1316 & 0.1265 & 0.0386 \\
chrF & 0.0620 & 0.0720 & 0.0671 & 0.0605 \\
chrF++ & 16.4425 & 20.7898 & 14.9770 & 14.0163 \\
BERTScore & 0.6584 & 0.6866 & 0.6415 & 0.6205 \\
BartScore & -4.2792 & -4.1679 & -4.2412 & -4.3022 \\
BLEURT & 0.7457 & 0.6707 & 0.6279 & 0.6143 \\
PQA & 0.4838 & 0.6113 & 0.2062 & 0.1189 \\
\bottomrule
\end{tabular}
\label{metric-multi-general}
\end{table}

\begin{figure*}[htbp]
\centering 
\includegraphics[scale=0.5]{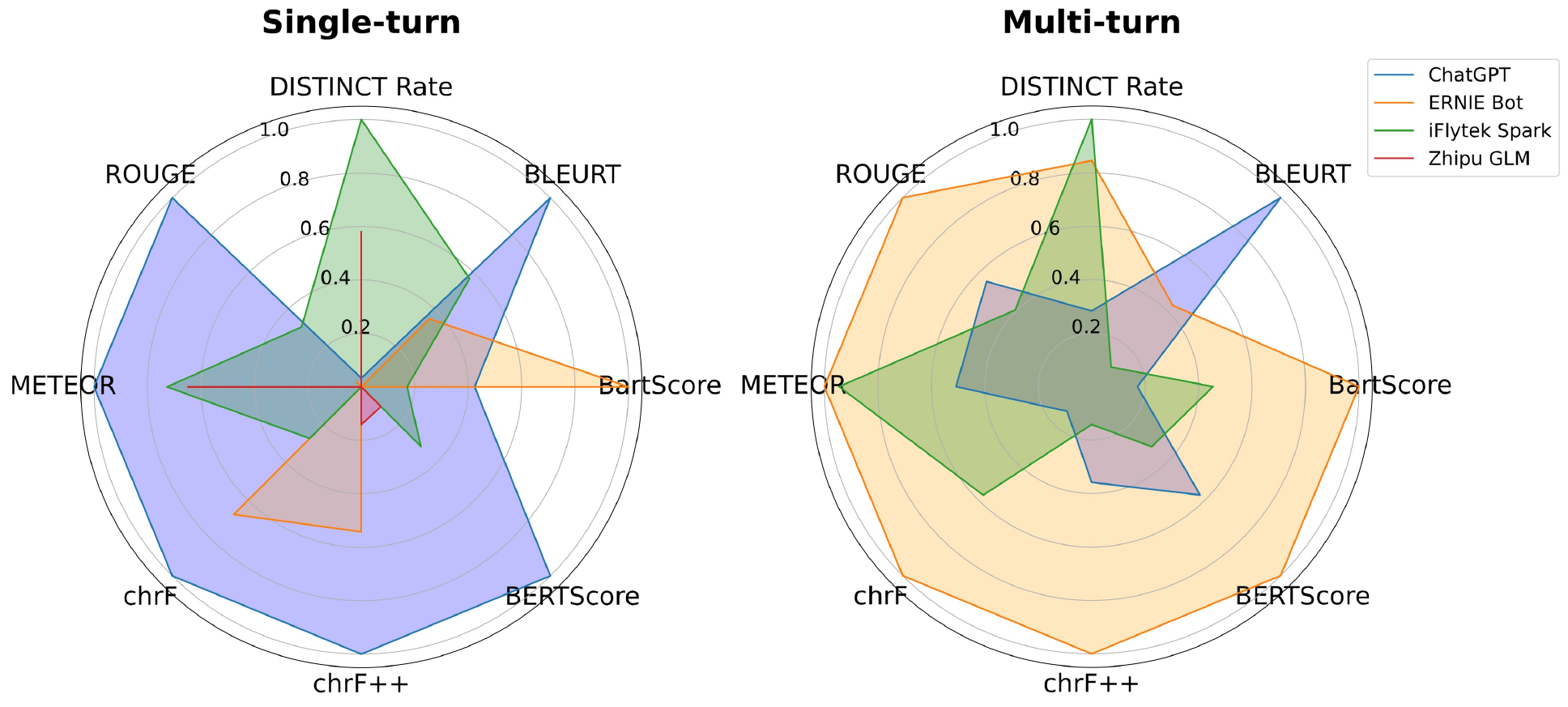} 
\caption{Radar charts of normalized evaluation metrics of general large language models.}
\label{radar-general}
\end{figure*}

In order to visually compare the performance of different language models in generating CBT dialogue, evaluation metrics of each model after normalization are further drawn as radar charts (Fig. \ref{radar-general}). In the case of single-turn conversations, ChatGPT is ahead of other models in the evaluation metrics of ROUGE, METEOR, chrF++, BERTScore and BLEURT. ROUGE, chrF and chrF++ mainly measure the degree of content overlap. In addition to the degree of content coincidence, METEOR also considers word order information, while BERTScore and BLEURT mainly measure semantic similarity. Therefore, compared with the other three large models, ChatGPT generates CBT text closer to the reference text, with more accurate and smooth semantic expression. On the other hand, iFlytek Spark has a higher DISTINCT Rate than other models, showing excellent diversity of text generation and less generated modular content. Although the BartScore also represents semantic similarity, the performance of ERNIE Bot on other metrics to measure semantic similarity is limited. Therefore, ERNIE Bot's ability to generate similar CBT conversation text and reference text is not as good as ChatGPT. In combination with the specific values of metrics in Table \ref{metric-single-general}, ChatGPT has the most outstanding performance in ROUGE, METEOR, chrF, CHRF ++, BERTScore and BLEURT, which are 22.51\%, 12.30\%, 6.21\%, 6.81\%, 2.71\% and 9.02\% higher than the second respectively, while the BartScore and DISTINCT Rate are slightly lower than the performance of ERNIE Bot and iFlytek Spark respectively by 1.24\% and 6.16\%. Comprehensive analysis shows that ChatGPT has the most outstanding ability to generate a single-turn CBT dialogue among the four LLMs.

In the case of multi-turn conversations, the DISTINCT Rate of iFlytek Spark is still outstanding, which is 1.52\% higher than the second place, and still maintains a certain advantage in text diversity. ERNIE Bot's ability to generate multi-turn CBT text is significantly improved, performing the best among the four language models in terms of metrics except DISTINCT Rate and BLEURT. Combined with metrics in Table \ref{metric-multi-general}, ERNIE Bot demonstrates superior performance in multi-turn conversations, achieving higher scores in ROUGE, METEOR, chrF, CHRF ++, BERTScore and BartScore by 32.37\%, 4.03\%, 7.30\%, 26.44\%, 4.28\% and 1.73\% respectively compared to the second model. This reflects ERNIE Bot's exceptional capability in conducting prolonged dialogues and retaining information. Analysis indicates that among the four language models, ERNIE Bot excels in generating multi-turn CBT text.

As shown in Table \ref{table-pqa-general}, under single-turn conversations, ChatGPT has the highest PQA, far exceeding the performance of the other LLMs, which are 123.44\%, 218.13\% and 333.65\% higher than ERNIE Bot, Zhipu GLM and iFlytek Spark respectively. This reflects that ChatGPT has a relatively mature ability to proactively inquire in a single-turn conversation, and is able to accurately understand the input and raise appropriate questions according to the input content, which also confirms the above mentioned excellent performance of ChatGPT in most universal evaluation metrics of LLMs. Under multi-turn conversations, the proactive inquiry performance of ERNIE Bot and iFlytek Spark has been improved, and ERNIE Bot's PQA is the highest among the four models, which were 26.37\%, 414.19\%, 196.46\% higher than ChatGPT, Zhipai GLM and iFlytek Spark respectively. This shows that ERNIE Bot's proactive inquiry ability in multi-turn conversations is the most prominent among the four models.

\begin{table}[htbp]
\centering
\normalsize
\setlength{\tabcolsep}{8pt}
\caption{PQA of general large language models under single-turn and multi-turn conversations}
\begin{tabular}{ccc}
\toprule
 & Single-turn PQA & Multi-turn PQA \\
\midrule
ChatGPT & 0.6713 & 0.4837 \\
ERNIE Bot & 0.3004 & 0.6112 \\
iFlytek Spark & 0.1548 & 0.2061 \\
Zhipu GLM & 0.2110 & 0.1188 \\
\bottomrule
\end{tabular}
\label{table-pqa-general}
\end{table}

\subsection{Knowledge Base Integrated Large Language Models}

After integrating knowledge base, generated CBT text example of each model under single-turn and multi-turn conversations are shown in Table \ref{tab-single-know} and Table \ref{tab-multi-know} respectively. In this case, the text generated in both single-turn and multi-turn conversations will be affected by the information contained in the CBT knowledge base. It obvious that LLMs with knowledge base generate more concise and realistic dialogue text when facing the same input of the patient, which is more in line with the dialogue style and pattern of a psychotherapist and a patient in real-life scenarios.

\begin{table}[htbp]
    \centering
    \renewcommand{\arraystretch}{1}
    \caption{Examples of generated single-turn CBT text of each knowledge base integrated large language model}
    \begin{tabular}{p{0.09\textwidth}p{0.32\textwidth}}
        \toprule
        \multicolumn{1}{c}{Name} & \multicolumn{1}{c}{Content} \\
        \midrule
        Input & Well, I really try to make sure I take at least one shower every day, but I often find that I haven't completed the other household chores that need to be done. There are many plates and things at home that I haven't cleaned up yet. \\
        \midrule
        Reference & It makes sense that when you have this feeling, you don't have the energy to do those things. Are there any other things you used to do frequently but don't do much now? \\
        \midrule
        ChatGPT & I know you mentioned that you had some difficulty coping with your emotional problems, not with your household chores. You'll panic and think about your chores. \\
        \midrule
        ERNIE Bot & I understand your problem. You might try making a chore list of all the tasks that need to be done, breaking them down into the smallest possible steps, and then doing them in smaller steps. Remember, by taking it one step at a time and not rushing it, you can slowly improve your household situation. \\
        \midrule
        iFlytek Spark & You also mention difficulties related to housework, such as showering but neglecting other chores, which may be because you feel tired or lack motivation. Depending on your description, you may need to take some steps to improve your situation. In addition, you may need to find a way to manage your anxiety, such as through meditation, deep breathing exercises, or seek professional counseling. \\
        \midrule
        Zhipu GLM & From the documentation you provided, you can see that there are some issues that can cause anxiety and behavior patterns. One suggestion worth considering is to work with a professional psychological counselor who can help you better understand these issues and provide personalized guidance and support. \\
        \bottomrule
    \end{tabular}
    \label{tab-single-know}
\end{table}

\begin{table}[htbp]
    \centering
    \renewcommand{\arraystretch}{1}
    \caption{Examples of generated multi-turn CBT text of each knowledge base integrated large language model}
    \begin{tabular}{p{0.09\textwidth}p{0.32\textwidth}}
        \toprule
        \multicolumn{1}{c}{Name} & \multicolumn{1}{c}{Content} \\
        \midrule
        Input & Well, I really try to make sure I take at least one shower every day, but I often find that I haven't completed the other household chores that need to be done. There are many plates and things at home that I haven't cleaned up yet. \\
        \midrule
        Reference & It makes sense that when you have this feeling, you don't have the energy to do those things. Are there any other things you used to do frequently but don't do much now? \\
        \midrule
        ChatGPT & These things are accumulating. So what else is there? \\
        \midrule
        ERNIE Bot & Consider finding some household help tools to lighten the load. Housework is an ongoing process, you don't have to rush to do it overnight, you will find that you can gradually improve the situation of housework. \\
        \midrule
        iFlytek Spark & I know your depressive mood. If you are willing, you can consider collaborating with mental health professionals who can help you identify and handle these behavioral patterns, and provide strategies to improve various aspects of your daily life. \\
        \midrule
        Zhipu GLM & From the information you have provided, many things do not work properly when a depressive mood occurs. I can see that you may have some problems in certain aspects of your life. \\
        \bottomrule
    \end{tabular}
    \label{tab-multi-know}
\end{table}

\begin{figure*}[htbp]
\centering 
\includegraphics[scale=0.25]{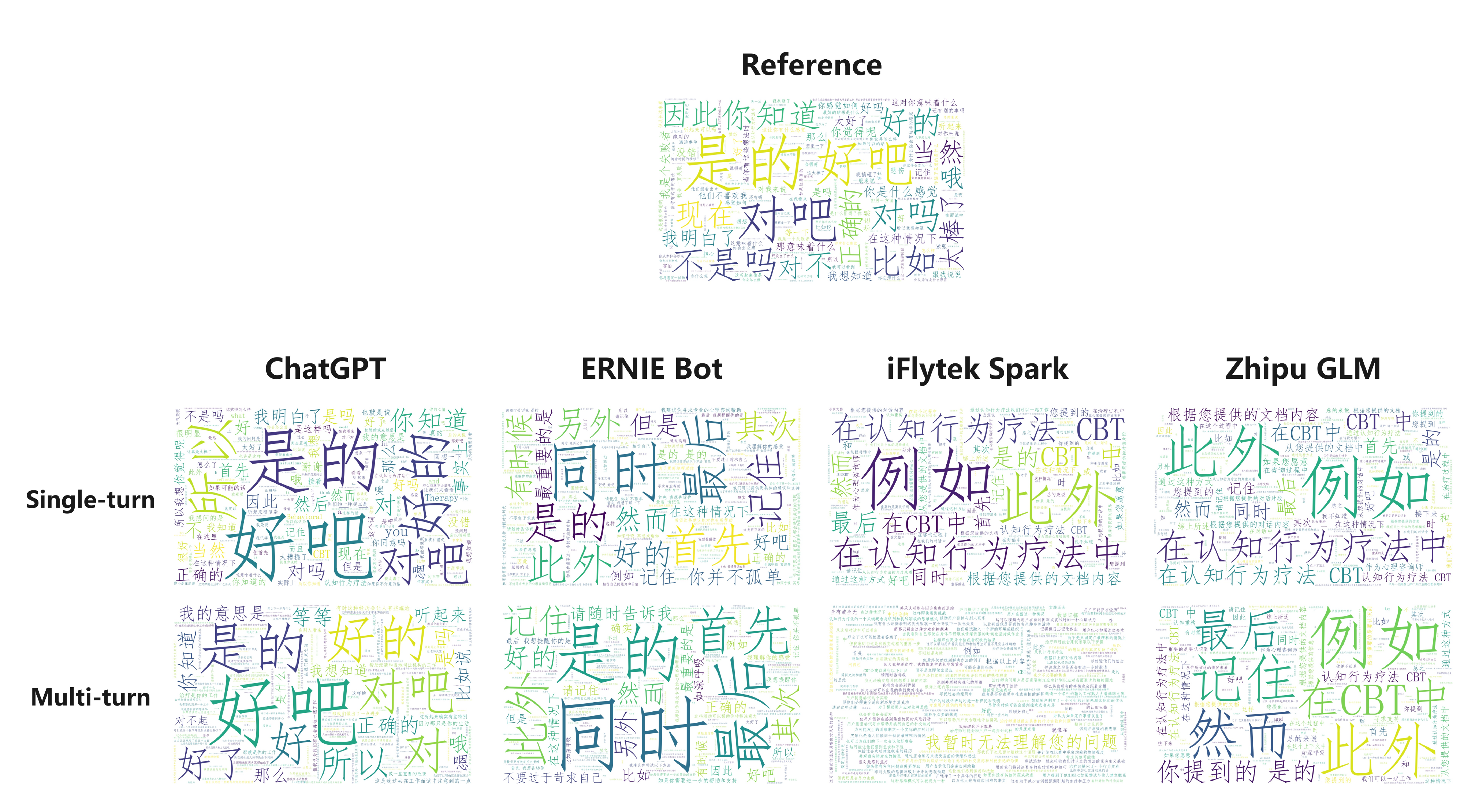} 
\caption{Cloud maps of the reference text and the text generated by knowledge base integrated large language models.}
\label{cloudmap-know}
\end{figure*}

As shown in Fig. \ref{cloudmap-know}, after integrating the knowledge base, both single-turn conversations and multi-turn conversations have their theme words with high frequency. Compared to Fig. \ref{cloudmap-general} without knowledge base integration, the difference between single-turn conversations and multi-turn conversations is significantly reduced, suggesting that most models have assimilated human psychotherapists' language style during real CBT processes from knowledge base. Meanwhile, model responses often display a colloquial characteristic, as evidenced by the frequent appearance of words such as "yeah", "right", "you know" and "but" in the word clouds of multiple LLMs. These all indicate that the integration of knowledge base provides a wealth of reference resources to enhance the capabilities of LLMs.

\begin{figure*}[htbp]
\centering 
\includegraphics[scale=0.4]{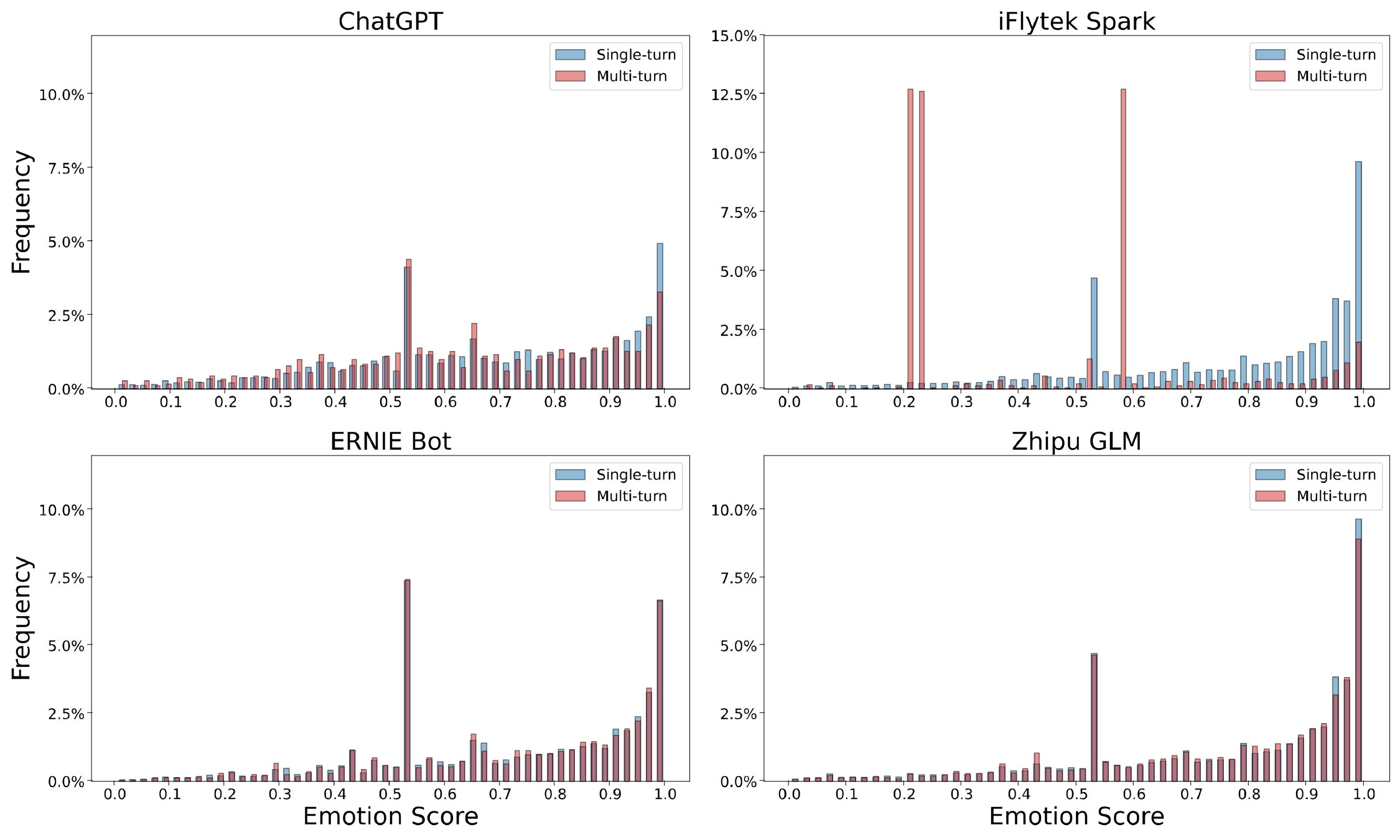} 
\caption{Emotion Scores of the CBT text generated by different knowledge base integrated large language models.}
\label{emotion-know}
\end{figure*}

The emotional tendency scores of each knowledge base integrated model under single-turn and multi-turn conversations are shown in Fig. \ref{emotion-know}. Whether conducting single-turn or multi-turn conversations, the emotional tendencies of CBT text generated by major language models are very positive, which helps improve the patient's psychological state and promotes longer interaction.

As depicted in Table \ref{metric-single-know}, under single-turn conversations, ROUGE, METEOR, chrF++, BERTScore and BLEURT of ChatGPT, ERNIE Bot and iFLYTEK Spark all increased, which indicates that the integrated knowledge base helps to improve the character, sentence and semantic similarity matching between the CBT text generated by models and the reference text to a certain extent. Because the text in the CBT knowledge base can provide more character level information, including proprietary vocabulary, phrases, sentence structures, etc., "automatic thought", "exposure therapy", etc. appear in CBT responses generated by LLMs. LLM can make use of the sentences in the knowledge base to perform semantic matching and correction on generated text, making the generated text more semantically consistent with the reference text, thereby making the generated text more closely related to the language style and expression of the reference text. However, the decrease of BartScore indicates that the fluency of generated CBT text is diminished after the integration of knowledge base. This could be attributed to the propensity of knowledge base to introduce additional external information, thereby increasing the likelihood of semantic incoherence or redundancy due to heightened information richness. When integrating these new knowledge, models frequently generate more verbose text, thus reducing the fluency and consistency of the text.

Interestingly, under multi-turn conversations, the four models did not demonstrate the same ability to integrate additional knowledge on most evaluation metrics. In terms of metrics other than chrF and BartScore, ChatGPT has a great improvement after integrating the knowledge base. And IFlytek Spark also shows significant enhancements in metrics other than DISTINCT Rate, METEOR and chrF++, showing strong knowledge base integration ability. However, most evaluation metrics of ERNIE Bot and Zhipu GLM are even lower after the integration of knowledge base.

\begin{table}[htbp]
\centering
\setlength{\tabcolsep}{3pt}
\caption{Evaluation metrics of knowledge base integrated large language models under single-turn conversations}
\begin{tabular}{ccccc}
\toprule
 & ChatGPT & ERNIE Bot & iFlytek Spark & Zhipu GLM \\
\midrule
DISTINCT Rate & 1.0493 & 0.9878 & 0.9760 & 0.8757 \\
ROUGE & 0.1246 & 0.1215 & 0.0913 & 0.0550 \\
METEOR & 0.1611 & 0.1308 & 0.1027 & 0.0226 \\
chrF & 0.0391 & 0.0779 & 0.06051 & 0.0543 \\
chrF++ & 16.6538 & 13.1096 & 14.5182 & 12.1826 \\
BERTScore & 0.6603 & 0.6492 & 0.6433 & 0.6047 \\
BartScore & -4.4765 & -4.4543 & -4.072 & -4.3194 \\
BLEURT & 0.7940 & 0.6769 & 0.6602 & 0.5194 \\
PQA & 0.6376 & 0.3007 & 0.2003 & 0.1510 \\
\bottomrule
\end{tabular}
\label{metric-single-know}
\end{table}

\begin{table}[htbp]
\centering
\setlength{\tabcolsep}{3pt}
\caption{Evaluation metrics of knowledge base integrated large language models under multi-turn conversations}
\begin{tabular}{ccccc}
\toprule
 & ChatGPT & ERNIE Bot & iFlytek Spark & Zhipu GLM \\
\midrule
DISTINCT Rate & 1.0846 & 0.9751 & 0.9216 & 0.8468 \\
ROUGE & 0.1467 & 0.1304 & 0.1145 & 0.0632 \\
METEOR & 0.1907 & 0.1288 & 0.1127 & 0.0242 \\
chrF & 0.0454 & 0.0816 & 0.0796 & 0.0586 \\
chrF++ & 17.9963 & 14.2157 & 14.9013 & 13.1691 \\
BERTScore & 0.6718 & 0.6574 & 0.6535 & 0.6175 \\
BartScore & -4.4246 & -4.4264 & -4.3119 & -4.2778 \\
BLEURT & 0.8459 & 0.7305 & 0.6558 & 0.6335 \\
PQA & 0.6126 & 0.2932 & 0.1949 & 0.1172 \\
\bottomrule
\end{tabular}
\label{metric-multi-know}
\end{table}

\begin{figure*}[htbp]
\centering 
\includegraphics[scale=0.5]{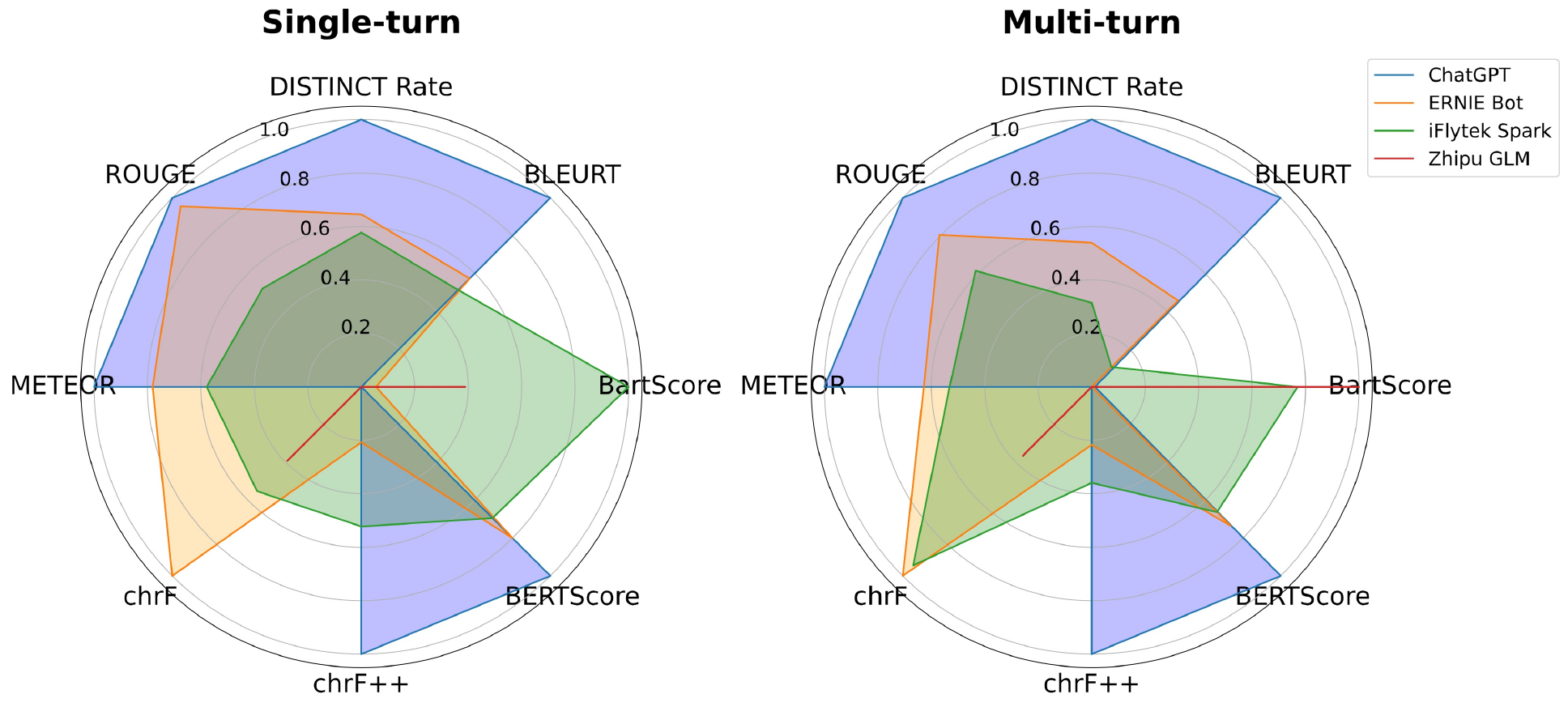} 
\caption{Radar charts of normalized evaluation metrics of knowledge base integrated large language models.}
\label{radar-know}
\end{figure*}

After presenting the normalized evaluation metrics using radar charts (Fig. \ref{radar-know}), it is evident that, whether in single-turn or multi-turn conversations, ChatGPT, integrated with a knowledge base, exhibits an absolute advantage over the other three large models in the majority of evaluation metrics. In single-turn conversations, ChatGPT has the best performance on DISTINCT Rate, ROUGE, METEOR, chrF++, BERTScore and BLEURT, exceeding the second by 6.23\%, 2.55\%, 23.17\%, 14.71\%, 1.71\% and 17.30\% respectively, while chrF and BartScore are 49.8\% and 9.93\% lower than ERNIE Bot and iFLYTEK Spark. In multi-turn conversations, ERNIE Bot still shows a certain advantage in chrF, exceeding the second-ranked iFlytek Spark by 2.51\%; Zhipu GLM maintains a weak advantage in BartScore, which is 0.8\% higher than the second ranked iFlytek Spark. In contrast, ChatGPT performs excellently in DISTINCT Rate, ROUGE, METR, chrF++, BERTScore and BLERT metrics, surpassing the second model by 11.23\%, 12.50\%, 48.06\%, 20.77\%, 2.19\%, and 15.80\%, highlighting the potential of ChatGPT for multi-turn conversations after integrating knowledge base.

After analyzing universal evaluation metrics, we further utilize PQA to assess and compare the proactively inquiry ability of different LLMs. As shown in Table \ref{table-pqa-know}, in single-turn conversations, ChatGPT has the highest PQA, far exceeding the performance of other language models, 112.10\%, 322.24\% and 218.24\% higher than ERNIE Bot, Zhipu GLM and iFLYTEK Spark, respectively, which reflects that ChatGPT has mature ability to proactively inquire during a single-turn conversation through the integration of knowledge base. In multi-turn conversations, the PQA of all LLM declined slightly. But the PQA of ChatGPT is still the highest among the four models, which is 108.97\%, 422.52\% and 214.40\% higher than ERNIE Bot, Zhipu GLM and iFLYTEK Spark respectively. This is consistent with ChatGPT's absolute advantage in most evaluation metrics.

\begin{table}[htbp]
\centering
\normalsize
\setlength{\tabcolsep}{8pt}
\caption{PQA of knowledge base integrated large language models under single-turn and multi-turn conversations}
\begin{tabular}{ccc}
\toprule
 & Single-turn PQA & Multi-turn PQA \\
\midrule
ChatGPT & 0.6713 & 0.4837 \\
ERNIE Bot & 0.3004 & 0.6112 \\
iFlytek Spark & 0.1548 & 0.2061 \\
Zhipu GLM & 0.2110 & 0.1188 \\
\bottomrule
\end{tabular}
\label{table-pqa-know}
\end{table}

\section{Conclusion} \label{Conclusion}
This paper explores the feasibility of using natural language processing technology to assist in psychological counseling and treatment by selecting four LLMs with outstanding dialogue generation question and answer capabilities to generate responses for CBT related single-turn and multi-turn dialogues. The ability of general LLMs to generate CBT dialogue text is compared in terms of text diversity, semantic similarity, text fluency, and active questioning ability. In addition, the ability of generating CBT dialogues of the knowledge base integrated LLMs is also tested and compared. The experimental result indicates that conducting CBT based on LLMs is feasible.

However, the experiment also has some limitations. Firstly, the limited number of CBT dialogues collected may not fully evaluate and test the LLM. Secondly, a limited knowledge base may not cover comprehensive knowledge of CBT, which can result in the responses generated by the model not providing sufficient and effective cognitive behavioral therapy guidance, thereby affecting treatment outcomes. Besides, CBT based on LLM involves the patient's privacy and personal information. Language models require access to the patient's conversation history and personal information to generate relevant responses, which may pose privacy and security risks and require careful handling.

\section*{Acknowledgment} \label{Acknowledgments}

This work was supported in part by the STI 2030-Major Projects (2021ZD0202002), in part by the National Natural Science Foundation of China (Grant No.62227807), in part by the Natural Science Foundation of Gansu Province,China(Grant No. 22JR5RA488), in part by the Fundamental Research Funds for the Central Universities(Grant No. lzujbky-2023-16). Supported by Supercomputing Center of Lanzhou University.

\balance

\clearpage
\section*{Appendix}
\begin{CJK}{UTF8}{gbsn}

\setcounter{table}{0}
\renewcommand{\thetable}{A\arabic{table}}

\begin{table}[H]
    \centering
    \renewcommand{\arraystretch}{1}
    \caption{Chinese examples of the CBT dataset}
    \begin{tabularx}{0.47\textwidth}{lX}
        \toprule
        \multicolumn{1}{c}{角色} & \multicolumn{1}{c}{内容} \\
        \midrule
        \multicolumn{2}{c}{情景：社交障碍} \\
        \midrule
        治疗师 & 你说你第一次感到焦虑是在五六年前，那时候你大概19岁吧? \\
        \midrule
        患者 & 是的。 \\
        \midrule
        治疗师 & 那时候发生了什么? \\
        \midrule
        患者 & 我在学校的时候一直有点害羞，不过我想这并没有像过去几年那样对事情产生太大影响。但我拒绝去参加派对，或者拒绝邀请。我认为我的工作能帮到我，因为我喜欢摄影，我真的很喜欢摄影，但是随着时间的推移，我的工作越来越少了。所以，这是我的错，我一直说不。 \\
        \midrule
        治疗师 & 在你的工作中，哪一种工作对你来说最困难? \\
        \midrule
        患者 & 我想当我必须和其他人一起工作的时候。我不喜欢那样。就像当我试图做一个活动或有很多人在那里的事情时，不只是摄影师，如果我一想到给我不认识的人拍照，我就会发现自己会拒绝，我是说，我想做，但太吓人了，所以我就不做了。 \\
        \midrule
        治疗师 & 当请你做一项工作，它会涉及到和一群人一起拍照，你需要有社交互动。你脑子里想的是什么? \\
        \midrule
        患者 & 我觉得我真的很热，出汗，我觉得每个人都在看着我，觉得我真的很蠢，觉得我不能完成我的工作，我就是个白痴。我的意思是，他们会认为我看起来像个怪人。 \\
        \bottomrule

        \multicolumn{2}{c}{情景：焦虑} \\
        \midrule
        患者 & 最近有挺多反省的事情，最近公司一直在加班也很忙，然后我们要做一个新的项目，这个项目就让我很头大，然后最近就是睡也睡不好。 \\
        \midrule
        治疗师 & 公司的一些项目需要去达成，对你造成非常多的困扰。 \\
        \midrule
        患者 & 对，今天我骑自行车来这里的路上，有个大卡车开过。我突然感到很紧张，不自觉地想象卡车翻车压到我会怎么样，让我很害怕。不知道是不是和之前工作上的经历有关。 \\
        \midrule
        治疗师 & 既然你先想到这样子的一个问题，我们干脆就先从一个卡车好像会压过来先来谈，过去有没有过这样子的一个想法在你脑中出现？ \\
        \midrule
        患者 & 之前的话也会有，偶尔有一些，然后当时觉得挺正常的，可能会人都怕死，但这次可能是最近这个压力比较大，可能今天来的，而且还是来咨询之前，然后这个感受会更强烈一些，就让我有点害怕了。\\
        \midrule
        治疗师 & 你觉得像卡车可能会翻过来，会压到你的可能性大概是多大？ \\
        \midrule
        患者 & 百分之十以下，感觉会比较低。 \\
        \bottomrule
    \end{tabularx}
    \label{chinese-dataset-example}
\end{table}

\begin{table}[H]
    \centering
    \renewcommand{\arraystretch}{1}
    \caption{Chinese examples of generated single-turn CBT text of each general large language model}
    \begin{tabular}{p{0.09\textwidth}p{0.32\textwidth}}
        \toprule
        \multicolumn{1}{c}{名称} & \multicolumn{1}{c}{内容} \\
        \midrule
        输入 & 嗯，我真的努力确保我每天至少洗个澡，但是其他需要做的家务，我经常发现他们没有完成。家里有很多盘子和东西我还没有清理干净。\\
        \midrule
        参考文本 & 当你有这种感觉时，你没有精力去做那些事情，这是有道理的。还有其他你以前经常做但现在不怎么做的事情吗? \\
        \midrule
        ChatGPT & 我想询问一下你觉得自己为什么能够坚持每天洗澡，但却觉得完成其他家务任务困难？你认为这两者之间有什么不同？ \\
        \midrule
        文心一言 & 家务事确实需要一定的时间和努力来保持家里的整洁和有序。如果你发现自己仍然无法完成家务事，可以考虑寻求家人或朋友的帮助，或者考虑雇佣家政服务来帮助你分担家务负担。 \\
        \midrule
        讯飞星火 & 这听起来像你可能需要更好地管理你的时间和家务任务。以下是一些可能有帮助的建议：制定计划，分解任务，设定优先级。记住，每个人都有忙碌的时候，有时候你可能无法完成所有的家务，这是正常的，不要对自己太苛刻。 \\
        \midrule
        智谱GLM & 家务活确实是一件既必要又繁琐的事情，特别是在忙碌或者生活节奏快的环境中。记住，最重要的是找到适合自己的生活节奏，保持家务和自我照顾之间的平衡，不要因为家务而忽略了自己的健康和幸福。 \\
        \bottomrule
    \end{tabular}
    \label{chinese-tab-single}
\end{table}

\begin{table}[htbp]
    \centering
    \renewcommand{\arraystretch}{1}
    \caption{Chinese examples of generated multi-turn CBT text of each general large language model}
    \begin{tabular}{p{0.09\textwidth}p{0.32\textwidth}}
        \toprule
        \multicolumn{1}{c}{名称} & \multicolumn{1}{c}{内容} \\
        \midrule
        输入 & 嗯，我真的努力确保我每天至少洗个澡，但是其他需要做的家务，我经常发现他们没有完成。家里有很多盘子和东西我还没有清理干净。 \\
        \midrule
        参考文本 & 当你有这种感觉时，你没有精力去做那些事情，这是有道理的。还有其他你以前经常做但现在不怎么做的事情吗? \\
        \midrule
        ChatGPT & 听起来你在日常生活中的自我照顾方面也有困难。那么，除了这些日常任务之外，还有什么其他的挑战是抑郁症带来的呢？ \\
        \midrule
        文心一言 & 听起来你已经尽力了，这很好。那么你有尝试找一些让你感到快乐或放松的活动吗? \\
        \midrule
        讯飞星火 & 抑郁会影响正常的生活习惯，有时候，药物治疗和心理治疗结合使用可以对缓解抑郁症状有所帮助。 \\
        \midrule
        智谱GLM & 抑郁症确实会改变一个人的日常习惯，如果你觉得情况严重，或者你的抑郁症状持续不断，强烈建议你咨询医生或心理健康专家。 \\
        \bottomrule
    \end{tabular}
    \label{chinese-tab-multi}
\end{table}

\begin{table}[htbp]
    \centering
    \renewcommand{\arraystretch}{1}
    \caption{Chinese examples of generated single-turn CBT text of each knowledge base integrated large language model}
    \begin{tabular}{p{0.09\textwidth}p{0.32\textwidth}}
        \toprule
        \multicolumn{1}{c}{名称} & \multicolumn{1}{c}{内容} \\
        \midrule
        输入 & 嗯，我真的努力确保我每天至少洗个澡，但是其他需要做的家务，我经常发现他们没有完成。家里有很多盘子和东西我还没有清理干净。 \\
        \midrule
        参考文本 & 当你有这种感觉时，你没有精力去做那些事情，这是有道理的。还有其他你以前经常做但现在不怎么做的事情吗? \\
        \midrule
        ChatGPT & 我知道你提到你在应对情绪问题方面，而不是在家务方面碰到了一些困难。你会恐慌起来然后联想到你的家务。 \\
        \midrule
        文心一言 & 很了解你的困扰。或许你可以尝试制定一个家务清单，将所有需要完成的任务一一列出，拆解成尽可能小的步骤，然后分步骤完成。记住，一步一步来，不要急于求成，你就可以慢慢改善家务状况。 \\
        \midrule
        讯飞星火 & 您还提到了与家务有关的困难，比如洗澡但忽略其他家务，这可能是因为您感到疲惫或缺乏动力。根据您的描述，您可能需要采取一些措施来改善您的状况。此外，您可能需要找到一种方法来管理您的焦虑，比如通过冥想、深呼吸练习或寻求专业的心理咨询。 \\
        \midrule
        智谱GLM & 从您提供的文档中，可以发现存在一些可能导致焦虑和行为模式的问题。一个值得考虑的建议是与专业的心理咨询师合作，可以帮助您更好地了解这些问题，并提供个性化的指导和支持。 \\
        \bottomrule
    \end{tabular}
    \label{chinese-tab-single-know}
\end{table}

\begin{table}[htbp]
    \centering
    \renewcommand{\arraystretch}{1}
    \caption{Chinese examples of generated multi-turn CBT text of each knowledge base integrated large language model}
    \begin{tabular}{p{0.09\textwidth}p{0.32\textwidth}}
        \toprule
        \multicolumn{1}{c}{名称} & \multicolumn{1}{c}{内容} \\
        \midrule
        输入 & 嗯，我真的努力确保我每天至少洗个澡，但是其他需要做的家务，我经常发现他们没有完成。家里有很多盘子和东西我还没有清理干净。 \\
        \midrule
        参考文本 & 当你有这种感觉时，你没有精力去做那些事情，这是有道理的。还有其他你以前经常做但现在不怎么做的事情吗? \\
        \midrule
        ChatGPT & 这些东西在积压。那么还有什么其他的事情呢？ \\
        \midrule
        文心一言 & 可以考虑寻找一些家务助手工具来减轻负担。家务是一个持续的过程，不必急于一蹴而就，你会发现能够逐渐改善家务状况。 \\
        \midrule
        讯飞星火 & 了解您的抑郁情绪。如果您愿意，可以考虑与心理健康专业人士合作，他们可以帮助您识别和处理这些行为模式，并提供策略来改善您的日常生活的各个方面。 \\
        \midrule
        智谱GLM & 从您提供的信息中，当抑郁情绪发作时，很多事情是没办法正常进行的。能看出您在生活中的某些方面可能存在一些问题。 \\
        \bottomrule
    \end{tabular}
    \label{chinese-tab-multi-know}
\end{table}

\end{CJK}
\end{document}